\documentclass{llncs}
\usepackage{srcltx}
\usepackage{amssymb}
\usepackage{bm}
\usepackage{subfigure}
\usepackage{amsmath}
\usepackage{amsbsy}
\usepackage{mathtools}
\usepackage{psfrag}
\usepackage{graphicx}
\usepackage{makeidx}

\newcommand{\citet}[1]{\cite{#1}}
\newcommand{\citep}[1]{\cite{#1}}

\usepackage{verbatim}
\usepackage{amssymb}
\usepackage{amsfonts}
\usepackage{bm}

\ifx\@waldernographics\@empty
\relax
\else
\fi%

\newcommand{\mat}[1]{\begin{pmatrix}#1\end{pmatrix}}

\newcommand{\psfragput}[4]{\psfrag{#1}{\begin{picture}(0,0)\put(#3,#4){#2}\end{picture}}}

\newcommand\tran{\top}

\newcommand{\ttran}{^{\tran}}

\newcommand{\diag}[1]{\text{diag}\lnb#1\rnb}
\newcommand{\mo}{^{-1}}

\newcommand{\chapternewpage}{~ \ifodd\value{page} \chapter*{} \fi}

\newcommand{\novelty}[3]{\ifthenelse{\equal{\value{#1}}{0}}{#3}{#2}\setcounter{#1}{1}}

\newcommand{\nn}{\nonumber}

\newcommand{\eg}{\textit{e.g.}}

\newcommand{\ie}{\textit{i.e.}}

\newcommand{\beqn}{\begin{equation}\/}
\newcommand{\eeqn}{\end{equation}\/}
\newcommand{\beqns}{\begin{equation*}\/}
\newcommand{\eeqns}{\end{equation*}\/}
\newcommand{\beqna}{\begin{eqnarray}\/}
\newcommand{\eeqna}{\end{eqnarray}\/}
\newcommand{\beqnas}{\begin{eqnarray*}\/}
\newcommand{\eeqnas}{\end{eqnarray*}\/}

\newcommand{\balign}{\begin{align}\/}
\newcommand{\ealign}{\end{align}\/}
\newcommand{\bals}{\begin{align*}\/}
\newcommand{\eals}{\end{align*}\/}

\providecommand{\norm}[1]{\lVert#1\rVert}
\newcommand{\abs}[1]{\left|#1\right|}
\newcommand{\normh}[1]{\left\|#1\right\|_{\Hcal}}
\newcommand{\mhalf}{^{-\frac{1}{2}}}
\newcommand{\half}{\frac{1}{2}}

\newcommand{\lnb}{\left(}
\newcommand{\rnb}{\right)}

\newcommand{\lcb}{\left\{}
\newcommand{\rcb}{\right\}}

\newcommand{\myvec}{\bm}

\newcommand{\alphavec}{\myvec{\alpha}}

\newcommand{\gammavec}{\myvec{\gamma}}

\newcommand{\etavec}{\myvec{\eta}}

\newcommand{\zerovec}{\myvec{0}}

\newcommand{\bvec}{\myvec{b}}

\newcommand{\evec}{\myvec{e}}
\newcommand{\fvec}{\myvec{f}}
\newcommand{\gvec}{\myvec{g}}

\newcommand{\rvec}{\myvec{r}}
\newcommand{\svec}{\myvec{s}}
\newcommand{\tvec}{\myvec{t}}

\newcommand{\xvec}{\myvec{x}}
\newcommand{\yvec}{\myvec{y}}
\newcommand{\zvec}{\myvec{z}}

\newcommand{\Hcal}{\mathcal{H}}

\newcommand{\Lcal}{\mathcal{L}}

\newcommand{\Ocal}{\mathcal{O}}

\newcommand{\Rcal}{\mathcal{R}}

\newcommand{\Xcal}{\mathcal{X}}

\newcommand{\realset}{\mathbb{R}}

  {\begin{center}\begin{Sbox}\begin{minipage}}%
  {\end{minipage}\end{Sbox}\fbox{\TheSbox}\end{center}}

  {\begin{center}\begin{Sbox}\begin{minipage}}%
  {\end{minipage}\end{Sbox}\Ovalbox{\TheSbox}\end{center}}

\newcommand{\cmcolor}{}

\newcommand{\var}[1]{\text{VAR}[#1]}
\newcommand{\varsub}[2]{\text{VAR}_{#2}[#1]}

\begin{document}

\title{Semi-Supervised Kernel PCA}
\author{Christian Walder, Ricardo Henao, Morten M{\o}rup and Lars Kai Hansen}
\institute{Informatics and Mathematical Modelling \\
Technical University of Denmark, DK-2800 \\
\email{\{chwa,rh,mm,lkh\}@imm.dtu.dk}}  

\maketitle

\begin{abstract}
We present three generalisations of Kernel Principal Components Analysis (KPCA) which incorporate knowledge of the class labels of a subset of the data points. 
The first, MV-KPCA, penalises within class variances similar to Fisher discriminant analysis. 
The second, LS-KPCA is a hybrid of least squares regression and kernel PCA. The final LR-KPCA is an iteratively reweighted version of the previous which achieves
a sigmoid loss function on the labeled points. We provide a theoretical risk bound as well as illustrative experiments on real and toy data sets.
\end{abstract}
\section{Introduction}
\label{SECintroduction}

\begin{figure}[t]
\begin{center}
   \psfragput{xlabel}{{\footnotesize{$f(\xvec)$}}}{-4}{-5}
   \psfragput{ylabel}{{\footnotesize{$\lnb 1 - \abs{f(\xvec)}\rnb_+$}}}{-22}{3}
   \includegraphics[width=7cm]{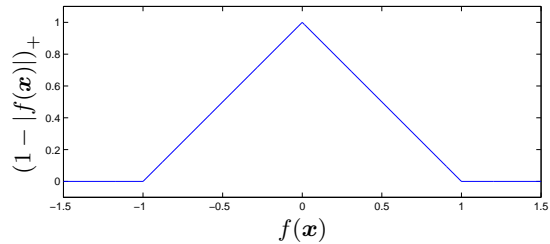}
\end{center}
\caption{\label{FIGtsvmloss}
The transductive SVM loss function for unlabeled points. Penalisation of this non-convex loss favours values $f(\xvec)$ which are either sufficiently positive or sufficiently negative, but not too close to zero.}
\end{figure}

In Semi-Supervised Learning (SSL) we are given a set of data points, only some of which come with class labels, and wish to infer a function which classifies new points. Alternatively we may not require the function but only its value on the unlabeled points, as in transduction. A considerable amount of work has recently been done here, see \eg\ \citep{zhu05survey,sslbook} for an overview and \citep{seegertechreport} for a discussion of the problem. Our approach is most closely related to the  class of discriminative algorithms exemplified by the transductive support vector machine or T-SVM \citep{vapnik98}. The classifying function of this natural semi-supervised extension of the (normal, or fully supervised) SVM can be written
\begin{align*}
f^\star = 
 \arg \min_{f\in\Hcal} \norm{f}^2_{\Hcal}
 + c_1 \sum_{i\in \Lcal} L(f(\xvec_i),t_i)
 + c_2 \sum_{j=1}^m U(f(\xvec_j)),
\end{align*}
where $\xvec_i\in\Xcal$ ($t_i\in\pm 1$) are the data points (labels), $\Lcal$ the indices of the labeled points, and $\Hcal$ a Reproducing Kernel Hilbert Space (RKHS). The labeled loss function proposed for the T-SVM is the usual hinge loss $L(f(\xvec),t)=(1-t f(\xvec))_{+}$ of the normal SVM, while the unlabeled loss $U(f(\xvec))=\lnb 1 - \abs{f(\xvec)}\rnb_+$ is the natural unlabeled analog of $L$, which we depict in Figure \ref{FIGtsvmloss}. Although it appears to be as sensible as the SVM, the non-convexity of $U$ makes the T-SVM much more difficult to handle, leading to various optimisation strategies \citep{chapsoptimisation}. 

In this paper we propose SSL algorithms which can be thought of either as generalisations of the (normally fully unsupervised) KPCA or as relaxations of the T-SVM in which $U$ takes the simpler form $U(f(\xvec))=-f(\xvec)^2$. Although this term is also non-convex (it is concave), it does lead to computational advantages. In particular, choosing also a quadratic loss for $L$ instead of the hinge, the problem is exactly solvable as we show in Section \ref{SEClskpca}. This combination of quadratic losses with exact solution is our least squares or LS-KPCA. Building on the useful exact solvability of this (still non-convex) proxy for T-SVM, we then propose as logistic regression or LR-KPCA, an iteratively reweighted version of LS-KPCA which gets closer to the T-SVM by implementing a sigmoidal loss function $L$, and utilising the exact solution of LS-KPCA in an inner loop.

\subsection{Overview and Organisation of the Paper}

We review KPCA in Section \ref{SECkpca} from a slightly unusual functional perspective. Our derivation relies on the representer theorem \citep{representer}, which turns out to make the discussion of our SSL generalisations of KPCA rather clean and straightforward. These generalisations of KPCA make up Section \ref{SECsskpca}. In Section \ref{SECmvkpca} we introduce MV-KPCA, which differs from KPCA in that the variance should be small over some prescribed subsets of the data. This is the simplest method we propose in that it is solved by a normal (generalised) eigenvalue problem. We argue in Section \ref{SECmvkpcadifficulties} that this formulation may be problematic. Addressing these problems, in Section \ref{SEClskpca} we introduce LS-KPCA, the method mentioned above with purely quadratic $L$ and $U$, which enjoys the risk bound we present in Section \ref{SECrisk}. LS-KPCA represents a greater departure from KPCA than MV-KPCA, but can also be solved exactly due to \citep{gandalf}. In Section \ref{SEClrkpca} we move further from KPCA, with an iterative reweighting scheme which utilises this exact solution in an inner loop in order to achieve a sigmoid rather than quadratic loss function, the intuition being that this may be more appropriate for classification problems.
A simple yet numerically stable optimisation procedure for LS- and LR-KPCA is outlined in Section \ref{SECsecular}.
In Section \ref{SECrelationship} we compare our algorithms to previous approaches, focussing on the Spectral Graph Transducer (SGT) of \citet{joachimssgt}.
We present results on standard benchmark data sets in Section \ref{SECexperiments}, and finish with some conclusions in Section \ref{SECconclusions}.

\section{Kernel PCA}
\label{SECkpca}

We treat KPCA \citep{nonlinearpca} slightly differently than usual, 
as the problem of finding
\begin{align}
\label{EQNkpca1}
f^\star = \arg \max_{f \in \Hcal} & ~ \sum_{i=1}^m \lnb f(\xvec_i)- \frac{1}{m} \sum_j f(\xvec_j) \rnb^2 \\
\label{EQNkpca2}
\text{subject to} & ~ \norm{f}_{\Hcal}^2 = 1,
\end{align}
where $\Hcal$ is the RKHS with kernel $k(\cdot,\cdot)$. The Lagrangian function associated with this problem is
\begin{align*}
L(f,\lambda) = \sum_{i=1}^m \lnb f(\xvec_i)-\frac{1}{m} \hspace{-1mm} \sum_j f(\xvec_j) \rnb^2 \hspace{-1.5mm} + \lambda\lnb \norm{f}_{\Hcal}^2  - 1\rnb.
\end{align*}
For the Lagrangian dual we maximise $L(f,\lambda)$ over $f\in\Hcal$. The representer theorem \citep{representer}, then implies 
$f^\star(\xvec) =\sum_{j=1}^m \alpha_i^\star k(\xvec_i,\xvec)$ for some $\alpha_1, \alpha_2, \ldots , \alpha_m \in \realset$. Combined with the reproducing property 
$f(\xvec) = \langle f , k(\xvec,\cdot)\rangle_{\Hcal}$, we obtain the simplification of \eqref{EQNkpca1} and \eqref{EQNkpca2} to 
\begin{align}
\label{EQNkpcaalpha1}
\alphavec^\star = \arg \max_{\alphavec \in \realset^m} & ~ \alphavec\ttran\lnb K\ttran K - K\ttran E_m K \rnb \alphavec \\
\label{EQNkpcaalpha2}
\text{subject to} & ~ \alphavec\ttran K \alphavec = 1,
\end{align}
where $K_{ij}=k(\xvec_i,\xvec_j)$ and $E_m$ is a square matrix of size $m$ with entries $\frac{1}{m}$, as it shall be throughout the paper. 
Imposing stationarity in $\alphavec$ on the Lagrangian of \eqref{EQNkpcaalpha1} and \eqref{EQNkpcaalpha2} we find that
$\alphavec^\star$ is the eigenvector $\alphavec$ with largest eigenvalue $\lambda$ 
of the generalised eigenvalue problem
\begin{align}
\label{EQNkpcaeigenproblemcentered}
K \alphavec = 
 \lambda \lnb K\ttran K - K\ttran E_m K \rnb \alphavec.
\end{align}

It is easy to verify that this formulation of KPCA is equivalent up to a re-normalisation to the original one \citep{nonlinearpca} in its \emph{centered} form. The simpler \emph{uncentered} version assumes that the data is centered in feature space. This leads to a slightly different eigenproblem formed by replacing the term
$\lnb K\ttran K - K\ttran E_m K \rnb$ with $K\ttran K$ in \eqref{EQNkpcaeigenproblemcentered}. Here we can recover the uncentered version of KPCA
by replacing the objective in \eqref{EQNkpca1} with $\sum_{i=1}^m f(\xvec_i)^2$. That is, the variance of the function values $f(\xvec_i)$ assuming they have zero mean.

\section{Semi Supervised Kernel PCA}
\label{SECsskpca}

We propose three means of incorporating label information into KPCA, with MV-KPCA (Section \ref{SECmvkpca}) incorporating a slightly different type of label information than LS- and LR-KPCA (Sections \ref{SEClskpca} and \ref{SEClrkpca}).

\subsection{Minimum Variance Kernel PCA (MV-KPCA)}
\label{SECmvkpca}

We begin with MV-KPCA,  which incorporates knowledge of pairwise, or rather group-wise similarity. This is the simplest method in that it is solved by an eigenproblem very similar to that of KPCA, and also in that it involves one extra parameter rather than two.
The idea of MV-KPCA is, reminiscent of the Kernel Fisher Discriminant \citep{kfd}, to modify the constraint \eqref{EQNkpca2} by adding a loss term based on the within-class variances. Given prescribed index sets $G_1, G_2, \ldots , G_l \subset \lcb 1, 2, \ldots m \rcb$ of similar elements from the data set $\xvec_1, \xvec_2, \ldots, \xvec_m$, the new constraint is
\begin{align*}
\norm{f}_{\Hcal}^2 + c \sum_{j=1}^l \sum_{i\in G_j} \lnb f(\xvec_i)- \frac{1}{\abs{G_j}} \sum_{j'\in G_j} f(\xvec_{j'}) \rnb^2 = 1,
\end{align*}
where $c\in\realset^+$ trades between KPCA for $c=0$ and increasing penalisation of within class variance for larger values of $c$. Note that (as in our experiments in this paper) the $G_j$ may be derived from categorical class labels by assigning points with the same label to the same group.
Once again we can apply the representer theorem to the Lagrangian of this problem. Since the augmented objective function is purely quadratic (like that of the original KPCA), its optimal solution is again found by an eigenproblem, this time
\begin{align}
\label{EQNmvkpcaeigenproblem}
\lnb K +  c\lnb \sum_i K_i\ttran K_i - K_i\ttran E_{\abs{G_i}} K_i \rnb \rnb \alphavec 
= \lambda \lnb K\ttran K -  K\ttran E_m K \rnb \alphavec,
\end{align}
where $K_i$ is the sub-matrix of $K$ taking rows $G_i$. Note that it is not possible to obtain a convex problem by replacing the
equality constraint with an inequality --- although the resulting problem is equivalent but on a convex feasible region,
in this case we would still be \emph{maximising} a convex function over that region.

\subsection{Difficulties Parameterising MV-KPCA}
\label{SECmvkpcadifficulties}

Since the objective function \eqref{EQNkpca1} and constraint \eqref{EQNkpca2} in KPCA (and MV-KPCA) both scale the same way (quadratically), 
the maximisation of one with the other fixed is equivalent to the maximisation of the ratio of the two. This is often referred to as the \textit{Rayleigh quotient} form of an eigenvalue problem \cite{hornyjohnson}. A consequence is that changing the constant on the right hand side of \eqref{EQNkpca2} only rescales $f^\star$, which could be problematic as we now argue.
The objective we are maximising is
\begin{align}
\label{EQNmvkpcaratio}
\max_{f\in\Hcal} \frac{\var{f}}{\normh{f}^2+c\sum_i\varsub{f}{i}},
\end{align}
where $\var{f}$ is the variance of the values of $f$ over all the $\xvec_i$ and $\varsub{f}{i}$ is the variance of the values of $f$ over the points indexed by $G_i$.
This looks like it may be interesting for SSL. After all, the objective function favours large variance on the unlabeled points 
while favouring small values of two fairly standard terms for regularisation and risk, namely an RKHS norm and a type of quadratic penalty. More importantly, although this may appear to be precisely the type of semi-supervised learning objective which tends to be hard to optimise due to its non-convexity, %
we can solve it in $\Ocal(m^3)$ time due to the convenient relationship with the eigenvalue problem \eqref{EQNmvkpcaeigenproblem}. The unfortunate part however, is that due to the fact that changing the constraint in \eqref{EQNkpca2} only multiplicatively scales 
the solution, there is no obvious way to trade between the numerator and the denominator of \eqref{EQNmvkpcaratio} in the same way we can trade off within the denominator
via the parameter $c$. This could be critical in SSL problems in which there are vastly different numbers of labeled and unlabeled points.
For a second multiplicative scaling parameter in the ratio \eqref{EQNmvkpcaratio} to be non-trivial however, 
at least one of the terms would need to scale non-quadratically.

\subsection{Least Squares Kernel PCA (LS-KPCA)}
\label{SEClskpca}

Continuing the previous argument, although many non purely quadratic surrogates for any of
the three terms in \eqref{EQNmvkpcaratio} are possible, few of the interesting ones will lead to 
computational problems as straightforward as solving the eigenvalue problem \eqref{EQNmvkpcaeigenproblem}. 
Fortunately however, the classic squared loss does turn out to be fairly convenient. Hence, as LS-KPCA we now propose the following modification of KPCA. First, instead of maximising the first term (the objective  \eqref{EQNkpca1}) with the second term (the constraint \eqref{EQNkpca2}) fixed, we minimise
the second with the first fixed. Actually, this is still KPCA as these problems are of course equivalent. Next, we add onto the new 
objective function a squared loss term, to get
\begin{align}
\label{EQNlskpca1}
f^\star = \arg\min_{f\in\Hcal} & ~ \normh{f}^2+ c \sum_{i\in\Lcal} \lnb f(\xvec_i) - y_i\rnb^2 \\
\label{EQNlskpca2}
\text{subject to} & ~ \var{f} = s^2.
\end{align}
An example solution of the above problem is depicted in Figure \ref{FIGlskpca}.
Unlike the original KPCA constraint and MV-KPCA, the above constraint does break the scale invariance of the ratio of the objective function and the constraint function and the problem cannot be written as a ratio similar to \eqref{EQNmvkpcaratio}. In other words, the part of \eqref{EQNlskpca1} which is linear in $f$ makes the relationship between $s$ 
and the corresponding optimal $f^\star$ non-trivial. Furthermore, although the parameterisation is unusual, we are now able to control the relative 
importance of the three terms, $\var{f}, \normh{f}^2$ and the squared error part of \eqref{EQNlskpca1}, via the parameters $c$ and $s^2$. 
Applying the representer theorem as before yields
\begin{align}
\label{EQNlskpcaalpha1}
\alphavec^\star = \arg \min_{\alphavec \in \realset^m} & ~ \alphavec\ttran K \alphavec + c \norm{K_{\Lcal}\alphavec-\tvec}^2 \\
\label{EQNlskpcaalpha2}
\text{subject to} & ~ \alphavec\ttran\lnb K\ttran K - K\ttran E_m K \rnb \alphavec = s^2,
\end{align}
where $\tvec \in \realset^{\abs{\Lcal}}$ is the 
sub-vector of $\yvec$ taking indices $\Lcal$, and $K_{\Lcal}$ is the submatrix of $K$ taking rows $\Lcal$.
\newcommand{\figwidth}{5.8cm}
\begin{figure}
\begin{center}
  \subfigure[KPCA, first eigenfunction]{
    \includegraphics[width=\figwidth]{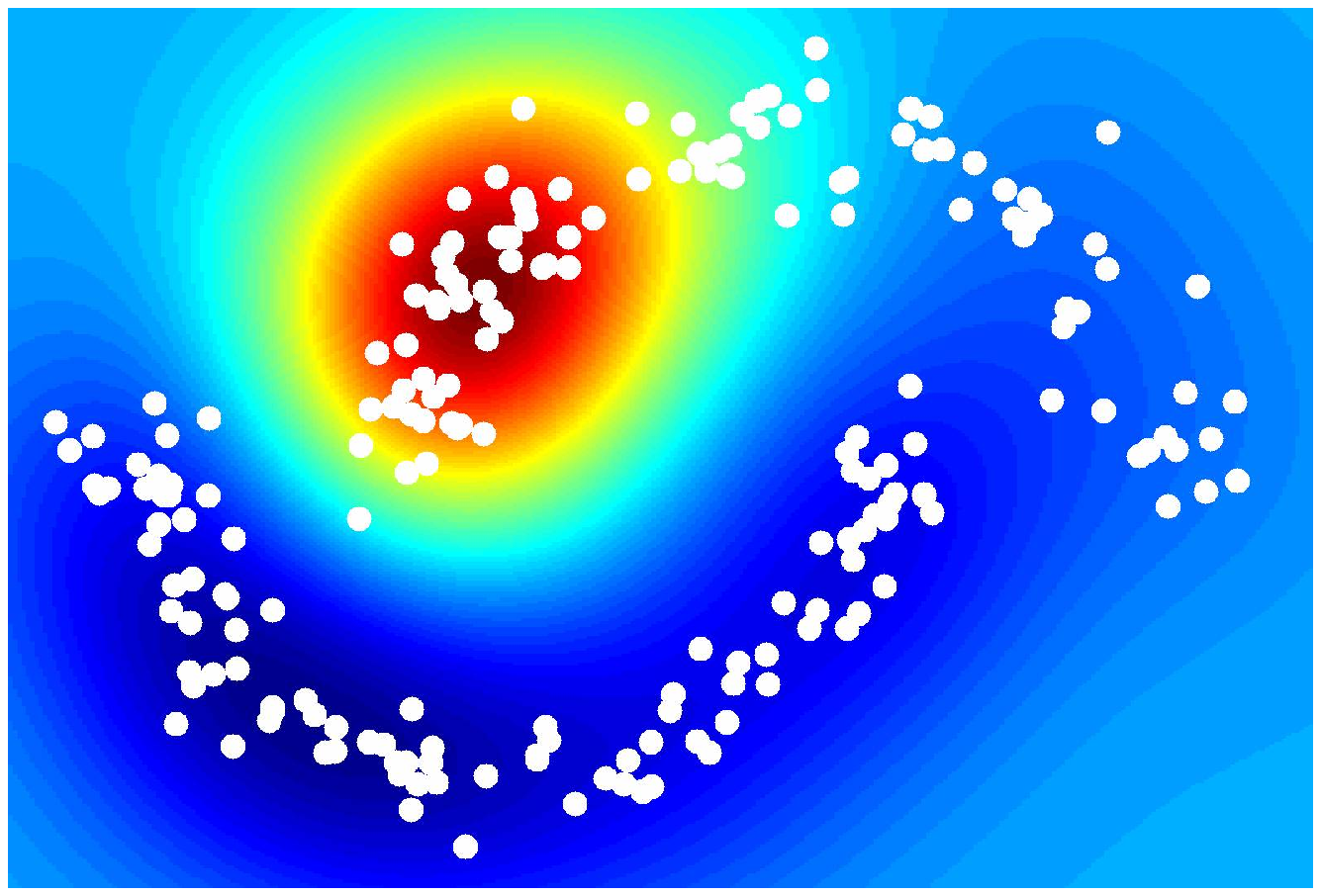}
  }
  \subfigure[KPCA, second eigenfunction]{
    \includegraphics[width=\figwidth]{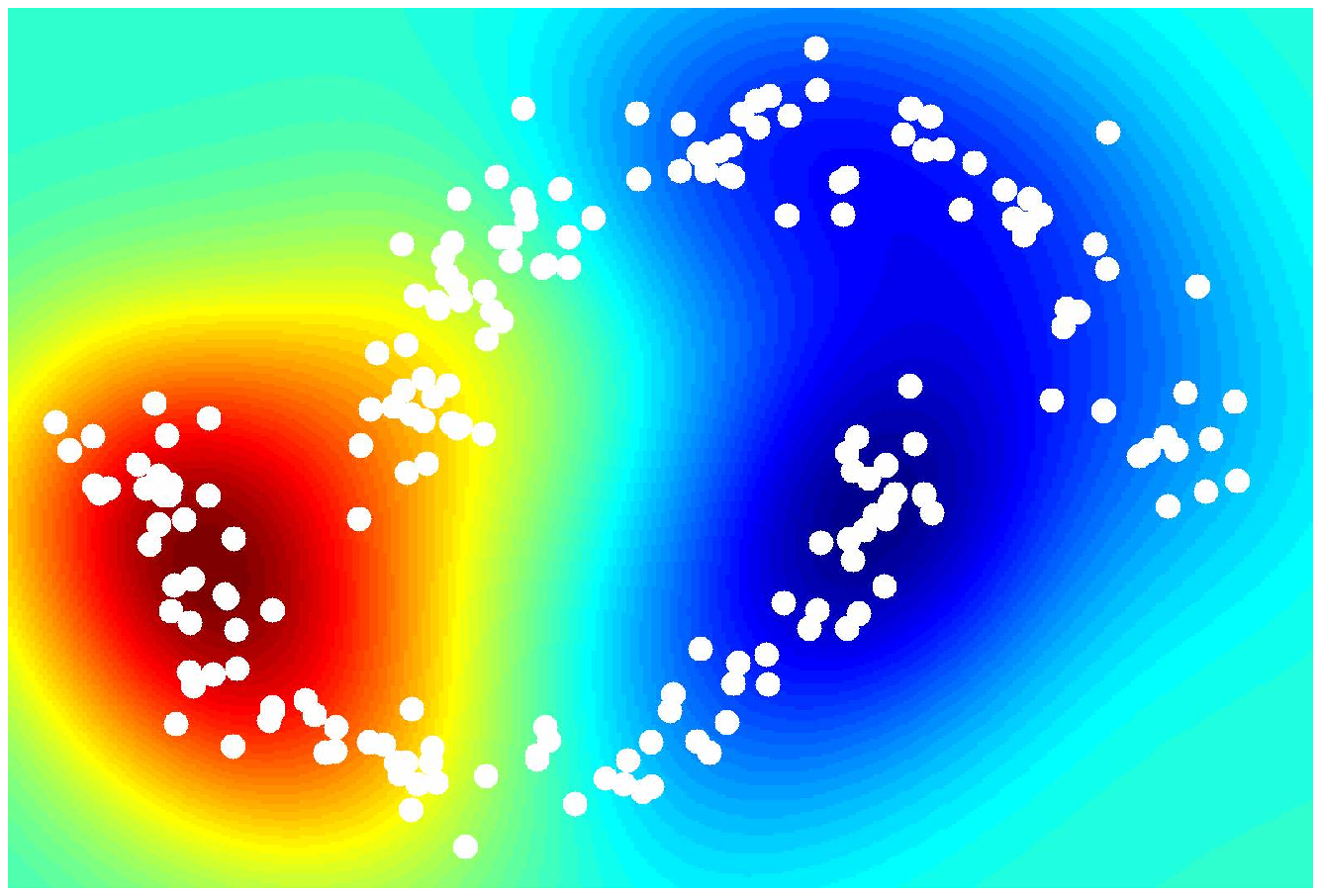}
  }
  \subfigure[KPCA, third eigenfunction]{
    \includegraphics[width=\figwidth]{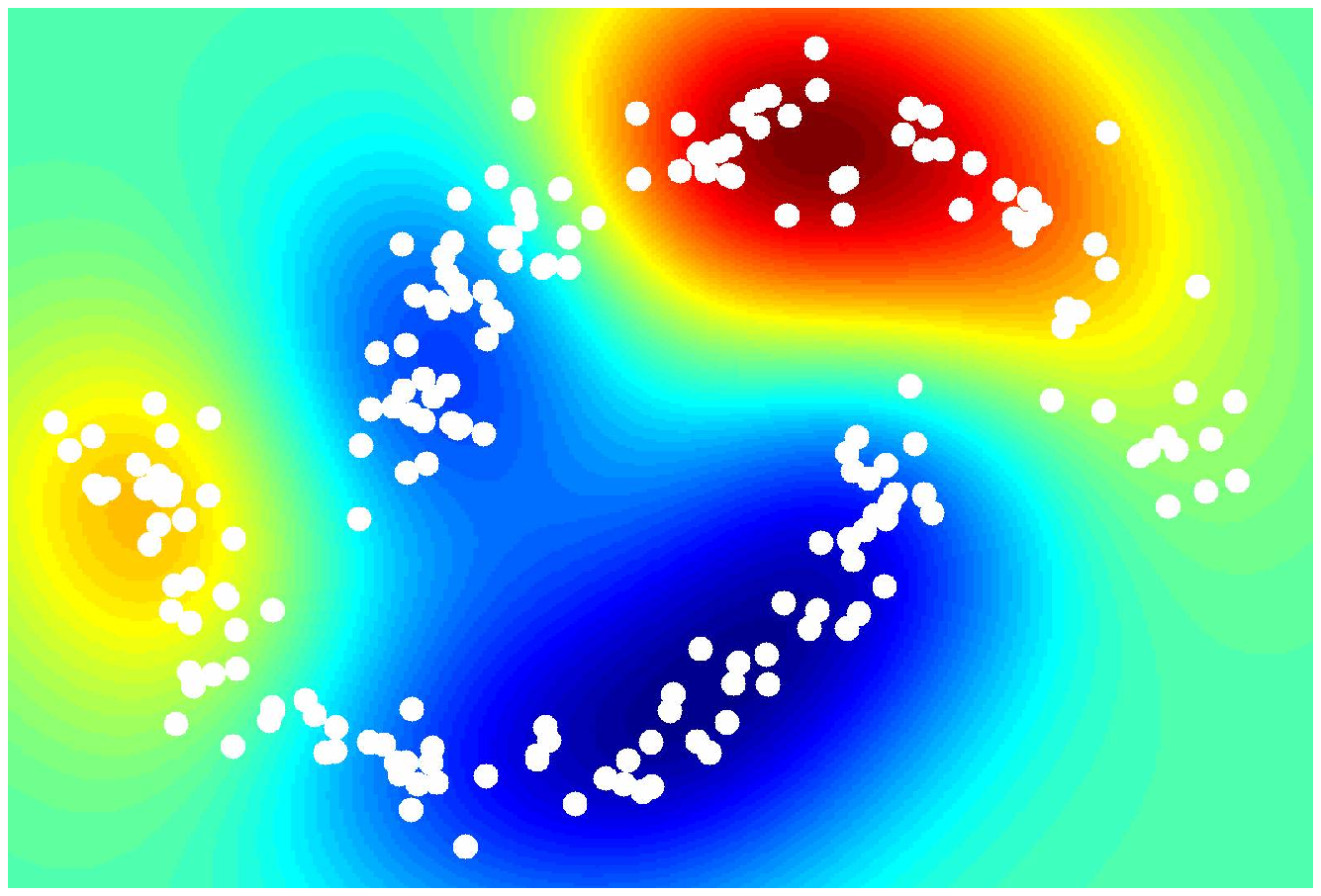}
  }
  \subfigure[LS-KPCA small $c$]{
    \includegraphics[width=\figwidth]{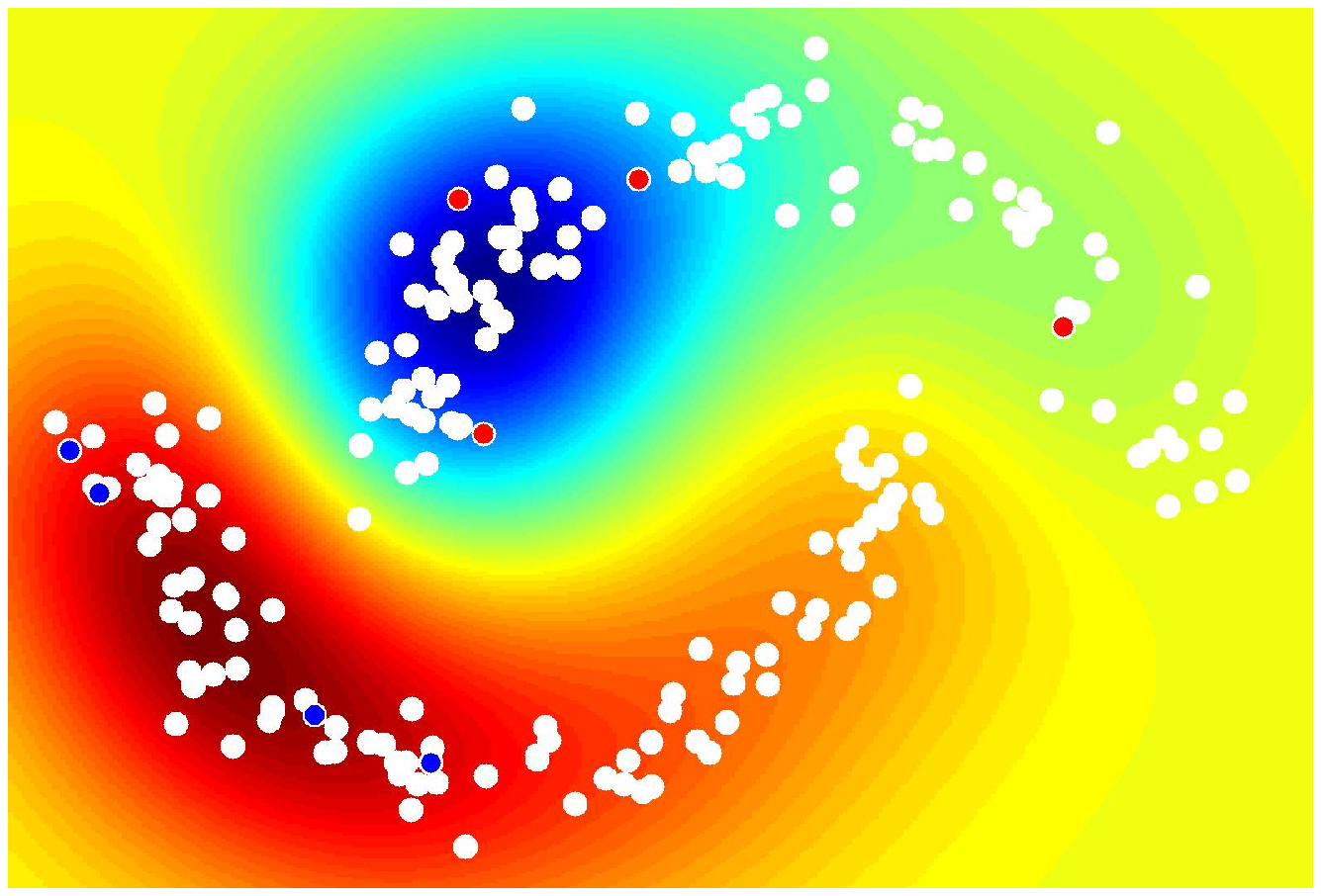}
  }
  \subfigure[LS-KPCA medium $c$]{
    \includegraphics[width=\figwidth]{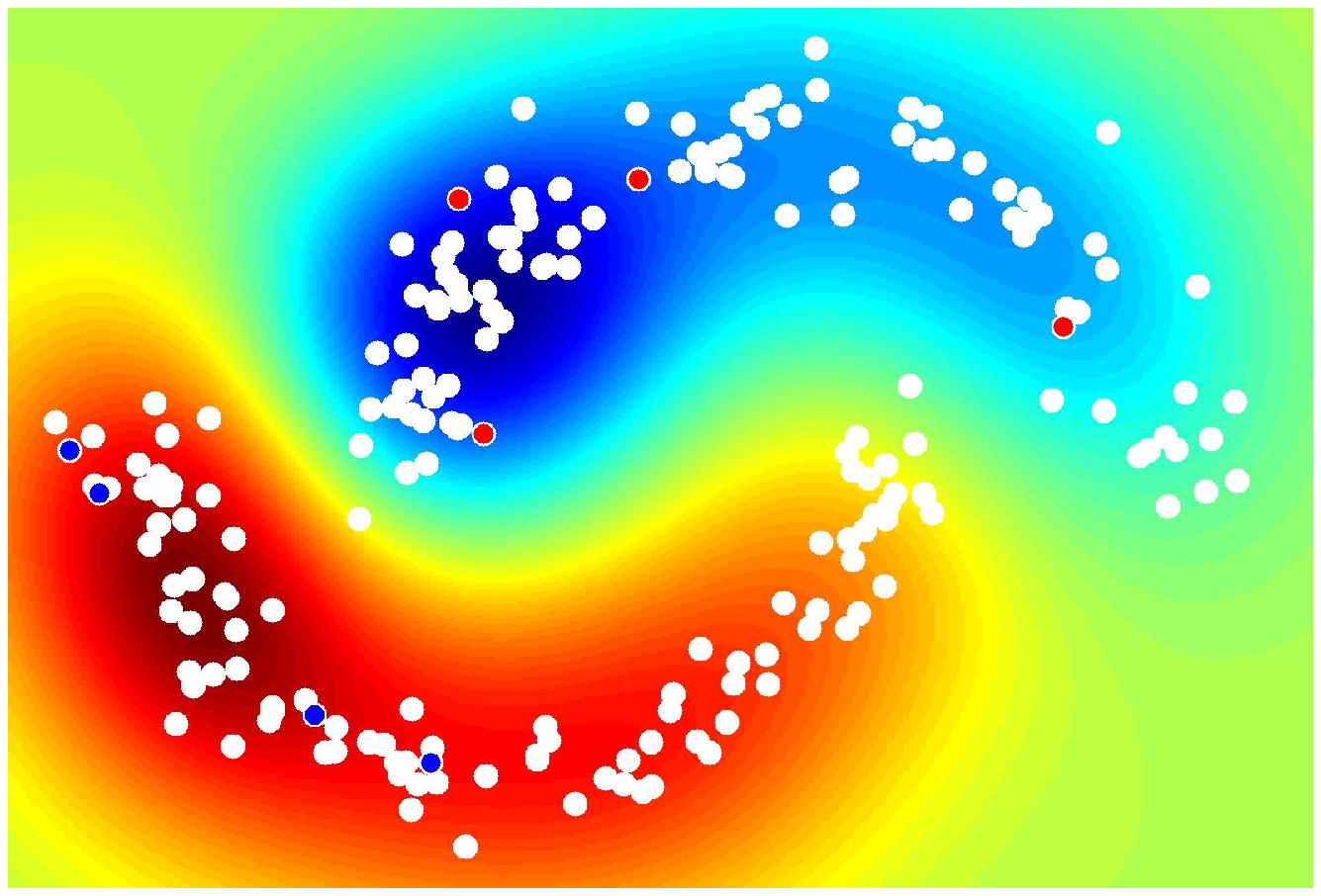}
  }
  \subfigure[LS-KPCA large $c$]{
    \includegraphics[width=\figwidth]{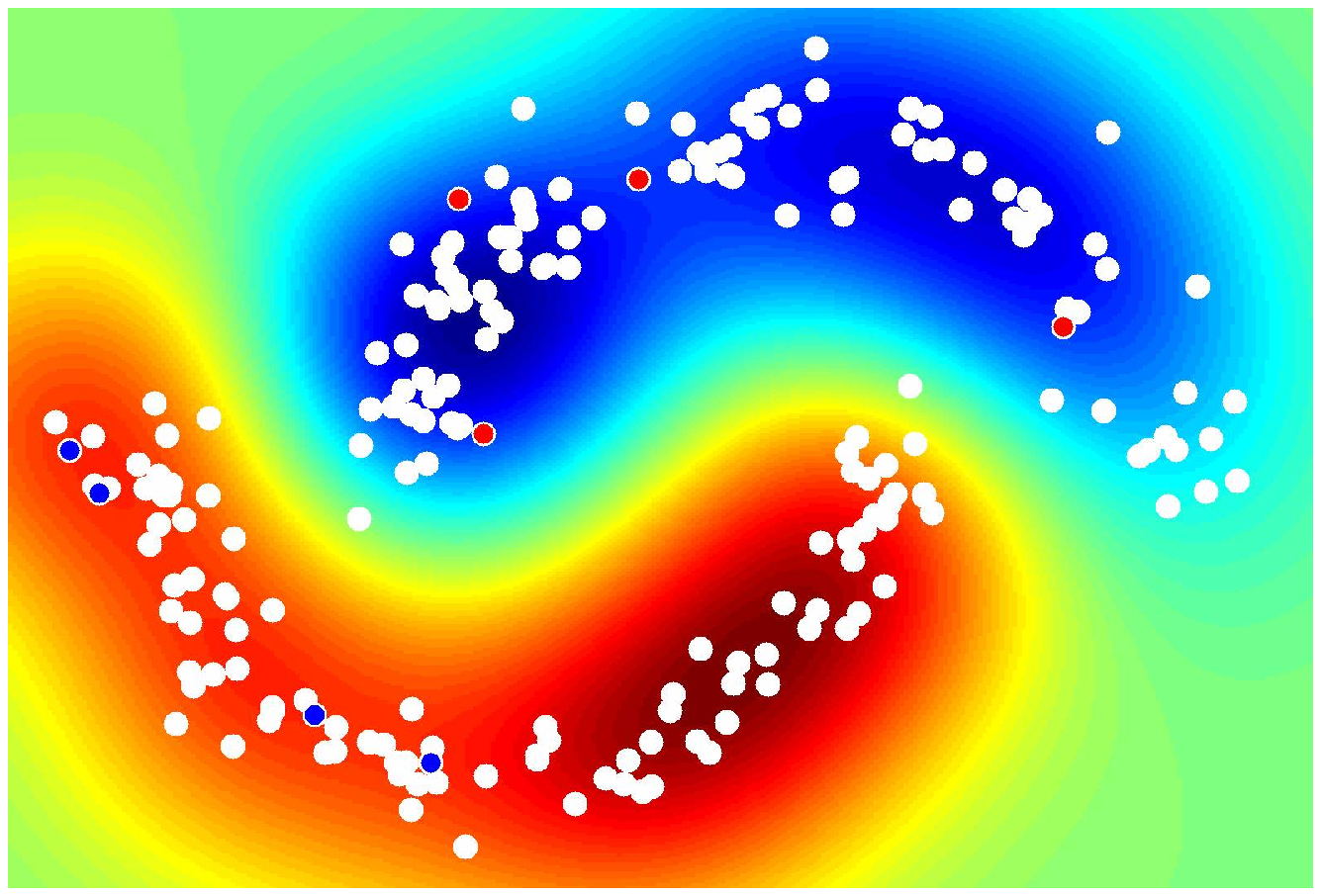}
  }
\end{center}
\caption{\label{FIGlskpca}
An example in $\realset^2$ of unlabeled (white) and labeled (red/blue, four per class) points using a spherical Gaussian kernel on the \textit{two moons} toy dataset. The value of $f^\star$ (from equation \eqref{EQNkpca1} for (a)-(c) and \eqref{EQNlskpca1} for (d)-(f)) is rendered in colour ranging from around +3 (dark red) to -3 (dark blue). LS-KPCA approaches KPCA for small $c$, hence there is a smooth transition (a)-(d)-(e)-(f), ignoring the sign change in (a) which is arbitrary for KPCA.  
}
\end{figure}
To solve \eqref{EQNlskpcaalpha1} and \eqref{EQNlskpcaalpha2} we can make use of the ideas in \citep{gandalf}, which studies the problem
\begin{align}
\label{EQNganderobjective}
\zvec^\star = \arg \min_{\zvec\in\realset^m} & ~ \zvec\ttran G \zvec - 2 \bvec\ttran \zvec \\
\label{EQNganderconstraint}
\text{subject to} & ~ \zvec\ttran\zvec=s^2.
\end{align}
It is shown %
that the solution to this non-convex problem is 
$\zvec^\star = \lnb G - \lambda^\star I\rnb\mo \bvec$, where $\lambda^\star$ is the smallest eigenvalue of the problem
\begin{align*}
\mat{ G & -I \\ -\frac{1}{s^2}\bvec\bvec\ttran & G} \mat{\gammavec \\ \etavec} = \lambda \mat{\gammavec \\ \etavec}.
\end{align*}
Note that this result was used in a related context \citep{joachimssgt}, as we discuss in Section \ref{SECrelationship}. 
Making the change of variables $\zvec=P^{\half} \alphavec$ where
\begin{align*}
P = K\ttran K - K\ttran E_m K,
\end{align*}
we can use this result to derive that 
\begin{align}
\label{EQNalphastar}
\alphavec^\star=\lnb C-\zeta^\star P\rnb\mo\bvec,
\end{align}
where
$C=K+cK_{\Lcal}\ttran K_{\Lcal}$, $\bvec = c K_{\Lcal} \tvec$
and $\zeta^\star$ is the smallest eigenvalue of 
the generalised eigenvalue problem
\begin{align}
\label{EQNalphastareig}
\mat{ C & - P \\ -\frac{1}{s^2}\bvec\bvec\ttran & C } \mat{\gammavec \\ \etavec} = \zeta \mat{P & \zerovec \\ \zerovec & P} \mat{\gammavec \\ \eta}.
\end{align}
The change of variables is unnecessary however, as we can repeat the arguments in \citep{gandalf} 
with the constraint in \eqref{EQNganderconstraint} replaced by $\zvec\ttran P \zvec = s^2$.

\subsection{Logistically Loss via Reweighting (LR-KPCA)}
\label{SEClrkpca}

Generalising LS-KPCA to arbitrary $L$ and $U$ for the labeled and unlabeled loss functions, we get
\begin{align}
\label{EQNlskpcarwf1}
f^\star = \arg\min_{f\in\Hcal} & ~ \normh{f}^2+c\sum_{i\in\Lcal} L\lnb f(\xvec_i),y_i \rnb \\
\label{EQNlskpcarwf2}
\text{subject to} & ~ \sum_i U (f(\xvec_i)) = s^2.
\end{align}
Note that the $U$ we intend here and for the remainder of the paper differs from that of the T-SVM formulation in Section \ref{SECintroduction} by a sign change. The purely quadratic losses of LS-KPA may not be appropriate for classification. Leaving $L$ and $U$ unspecified but abusing the notation by extending them element-wise to vectors, the representer theorem still applies, so we can write the problem in $\alphavec$ as
\begin{align}
\label{EQNlskpcarwa1}
\alphavec^\star = \arg\min_{\alphavec\in\realset^m} & ~ \alphavec\ttran K \alphavec+ c \evec\ttran L(K\alphavec,\yvec) \\
\label{EQNlskpcarwa2}
\text{subject to} & ~ \evec\ttran U (K\alphavec) = s^2,
\end{align}
where $\evec$ is a vector of ones. We would like to use more sophisticated losses $U$ and $L$ in the above formulation. For this, it is natural to try to leverage the powerful result that we can solve the least squares formulation exactly, by employing the \emph{iteratively reweighted least squares} idea \citep{nelder}. 
This is essentially a Newton-Raphson method, but with the interpretation that each step solves a least squares problem 
with modified weights.%

A Newton-Raphson step solves a local second order approximation of the problem.
To be able to apply the exact solution of the previous section, 
we are forced to choose a $U$ which is purely second order (\ie\ with no linear term). Then the local second order approximation of the constraint \eqref{EQNlskpcarwa2} is still purely second order (and exact), and the form of the optimisation problem remains that of LS-KPCA. Hence we maintain our initial choice $U(f(\xvec))=f(\xvec)^2$. We can try to improve on $L$ however, as doing so does not change the form of the objective function on taking a local second order approximation, since the LS-KPCA objective \eqref{EQNlskpcaalpha1} already has a  linear part. Hence, motivated by logistic regression, as LR-KPCA we propose the sigmoid  
\begin{align*}
L(f(\xvec),y)=1/(1+\exp(-y f(\xvec))
\end{align*}
as the loss term for labeled points.
A Huber-like differentiable approximation of the support vector machine hinge loss could just as easily be used, however.

By arguments similar to those of logistic regression \citep{nelder}, 
we can solve \eqref{EQNlskpcarwa1}-\eqref{EQNlskpcarwa2} by iteratively reweighted LS-KPCA. 
We can derive as usual that the following steps constitute a Newton-Raphson update. Given the current solution $\alphavec_n$, 
we compute $\gvec,\zvec,\svec \in \realset^{\abs{\Lcal}}$ as $\gvec =  K_{\Lcal} \alphavec_n$, and 
\begin{align}
\label{EQNrewaa}
z_i & = 1/\lnb 1+\exp( -t_i g_i) \rnb, \\
\label{EQNrewbb}
\nn r_i & = z_i(1-z_i), \\
\nn s_i & = g_i-(z_i-t_i) (1-z_i)/z_i,
\end{align}
for $i = 1,2,\ldots,\abs{\Lcal}$. 
The next iterate $\alphavec_{n+1}$ is defined like
$\alphavec^\star$ of \eqref{EQNalphastar} and \eqref{EQNalphastareig}, but with a different $C$ and $\bvec$, which now depend on $\rvec$ and $\svec$ according to
\begin{align*}
C  = K+cK_{\Lcal}\ttran R K_{\Lcal},\quad 
\bvec  = c K_{\Lcal}\ttran R \svec,
\end{align*}
where $R$ is diagonal with $R_{ii}=r_i$. 

Due to the form of the logistic function \eqref{EQNrewaa} we have that $r_i>0$ and so the resulting Hessian $C$ is always positive definite. As is also the case for normal logistic regression however, we have no guarantee that it will improve the objective function, making some form of back-tracking line search necessary. Due to the constraint \eqref{EQNlskpcarwf2}, this is not as simple as moving back on the line $\lambda \alpha_n + (1-\lambda)\alpha_{n-1}, 0 \leq \lambda \leq 1$. Instead,  in order to guarantee convergence we check the objective function, and as long as it is not better than the previous iterate, we solve a modified problem with an additional regularisation term $\lambda \norm{\alpha_n-\alpha_{n-1}}^2$, where $\lambda$ is a parameter we increase until we see an improvement in the (unmodified) objective function. %
It is important to note that similar line search heuristics are also required in the iteratively reweighted maximimum likelihood solver of the standard logistic regression model. %

\newcommand{\figwidthb}{5.7cm}
\begin{figure}[t]
\begin{center}
  \subfigure[LS-KPCA]{
    \includegraphics[width=\figwidthb]{figs/ls.eps}
  }
~
  \subfigure[LR-KPCA]{
    \includegraphics[width=\figwidthb]{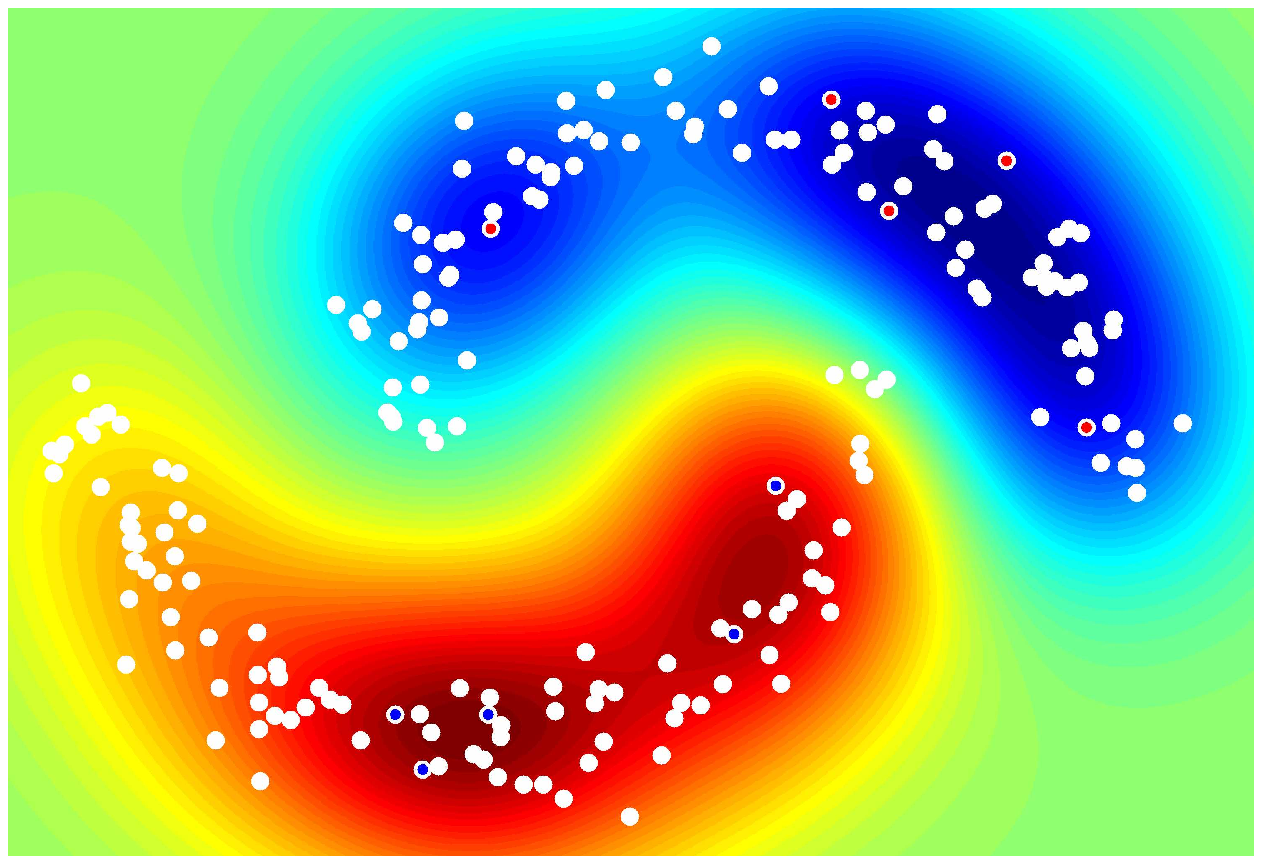}
  }
\end{center}
\caption{\label{FIGlrkpca}
LS-KPCA (\emph{left}) suffers here due to the value of $s^2$ in \eqref{EQNlskpcaalpha2} being large. This essentially enforces a large squared value of the function at certain points, conflicting with the squared loss which favours values near $\pm 1$, so that  the energy of the function gets concentrated away from the labeled points. The sigmoid loss of LR-KPCA (\emph{right}) can discount labeled points which are well classified, leaving the energy unhampered. The values range from around -50 (dark red) to +10 (dark blue) for LS-KPCA, and from around -3 to +3 for LR-KPCA.}
\end{figure}

\section{Efficient Solution}
\label{SECsecular}

We need to compute the $\zeta^\star$ of \eqref{EQNalphastar}, both for LS-KPCA and the inner loop of LR-KPCA.  As argued in \citep{gandalf}, doing so via the eigenvalue equation \eqref{EQNalphastareig} directly can be highly unstable since the matrix on the left hand side is not symmetric, is twice the size of the normal KPCA eigenvalue problem, and is typically badly conditioned. 
From \eqref{EQNlskpcaalpha2}, the solution $\alphavec^\star$ must satisfy $\alphavec^\top P\alphavec=s^2$. As it was shown in \citep{gandalf}, for $\zeta<\delta$ where $\delta$ is the smallest eigenvalue of $C$, \eqref{EQNganderobjective} and \eqref{EQNganderconstraint} have a unique solution if and only if the characteristic polynomial (or secular equation)  $f(\zeta)=\alphavec^\top P\alphavec-s^2=0$ is satisfied, provided that $f(\zeta)$ is strictly increasing for $\zeta\in(-\infty,\delta)$. As a result, to obtain $\alphavec^\star$ we need to find the unique root of $f(\zeta)$ to the left of $\delta-|{\bf u}^\top {\bf b}|/s$, where ${\bf u}$ is the eigenvector with associated eigenvalue $\delta$. Here we use Brent's method \citep{brentsmethod}, allowing $\zeta^\star$ to be calculated to high precision. Note that the uncentered $\var{f}$  allows us to make a straight-forward change of variables $\fvec = K \alphavec$ and solve the problem in $\fvec$. The resulting eigenvalue problem is of the form \eqref{EQNganderobjective}-\eqref{EQNganderconstraint}, with an isotropic constraint, and can be solved more efficiently via (to use their terminology) the simpler \emph{explicit} characteristic polynomial of \citep{gandalf}, as opposed to the \emph{implicit} one used here. Transforming to an isotropically constrained variable in this way would be possible for the centered formulation as well, however this transformation leads to numerical problems due to the required matrix inverse. Moreover, the inverse in \eqref{EQNalphastar} means that it is not possible to reduce the computational time complexity from cubic.

\section{Risk Bound}
\label{SECrisk}

It is straightforward to obtain a risk bound for LS-KPCA using \citep{pechyony}, with the analysis being similar to that of the SGT in that work.
Assume the uncentered version of \eqref{EQNlskpca2}, so that $P=K\ttran K$. From \eqref{EQNlskpcaalpha1} and \eqref{EQNlskpcaalpha2} we have that the soft decision function values $\zvec^\star=K\alphavec^\star$, an Unlabeled-Labeled Decomposition (ULD). %
Letting $m=l+n$, where $l$ (resp. $n$) is the number of labeled (unlabeled) points, we can obtain an error bound for LS-KPCA

\begin{theorem}[\citep{pechyony}]

Let $\zvec^\star=K\alphavec^\star$ be the ULD of a transductive algorithm. Let $\|\alphavec\|_2\leq \mu$, $c=\sqrt{32\ln(4e)/3}<5.05$ and $r=1/l+1/n$. %
With probability of at least $1-\delta$ over the choice of the training set ${\cal L}$ from $X$, for all $\zvec$ in the set of all possible hypothesis that can be generated by the algorithm for a given $X$, all possible partitions and all possible labelings of the training set, the risk $\Rcal_n(\zvec)$ is bounded from above by
\begin{align*}
{\cal R}_l(\zvec) +\sqrt{\frac{2\mu^2}{ln}\|K\|^2_{\mathrm{ Fro}}}+c r\sqrt{\min(l,n)}+\sqrt{2r\ln\frac{1}{\delta}} \ .
\end{align*}
\end{theorem}

To apply this to LS-KPCA, we replace $\zvec^\star=K\alphavec^\star$ in \eqref{EQNlskpcaalpha1} and \eqref{EQNlskpcaalpha2}, eigen-decompose $G = K\mo C K\mo=QDQ^\top$ (so that $Q\ttran Q = I$) and put $\alphavec=Q^\top {\bf z}$. We see ${\bf z}^\star=Q\alphavec^\star$ with $\alphavec^\star$ as in (16). Since $Q$ depends only on the data ${\bf X}$, the latter is a ULD of LS-KPCA. From (12) $\alphavec^\top\alphavec=s^2$ and $\|Q\|^2_{\mathrm{ Fro}}=q$, where $q$ is the rank of $G$. Hence ${\cal R}_n({\bf z})$ is bounded from above by
\begin{align*}
	{\cal R}_l({\bf z})+\sqrt{\frac{2q s^2}{ln}}+c r\sqrt{\min(l,n)}+\sqrt{2r\ln\frac{1}{\delta}} \ . 
\end{align*}

The second term is an upper bound on the Rademacher complexity of ULD algorithms, for the SGT it is $\sqrt{2\widetilde{q} r}$ \citep{pechyony}, where $\widetilde{q}$ is the number of non-zero eigenvalues of the Laplacian. 
Since both algorithms use a squared loss for $\Rcal_l$, their bounds differ only by the Rademacher complexity, which may even be made equal by choosing $s$ appropriately. This is to be expected, since as we explain in the next section LS-KPCA differs from the SGT only by its regulariser.

\section{Relationship to Other Methods}
\label{SECrelationship}

Firstly, MV-KPCA is similar to the Kernel Fisher Discriminant, penalising a different set of variances but also leading to an eigenvalue problem \citep{kfd}. Next, LS-KPCA is related to \citep{joachimssgt} and \citep{vapnikvolume}, the former relationship being the clearest.
In particular, following \citep{graphlaplacianrkhs} we can interpret Joachims' SGT \citep{joachimssgt} as a special case of LS-KPCA. To do this, in \eqref{EQNlskpca1} and \eqref{EQNlskpca2} we define the RKHS of functions as the set of real valued functions defined on the vertices of the graph and satisfying a particular linear constraint (normalised cut balancing constraint, related to our centered variance for \eqref{EQNlskpca2}) so that $\Hcal = \lcb \fvec \in \realset^m : \fvec\ttran\evec=0\rcb$ where $\evec$ is a vector of ones. We define the graph Laplacian matrix $L$ as in \citep{joachimssgt}, and let the kernel matrix $K$ be given by $L^{+}$, the pseudo-inverse of $L$. If we further restrict to the simpler uncentered version of \eqref{EQNlskpca2}, and use the fact that the first eigenvector of $L$ is a scaled $\evec$, then simplifications lead to equations (19) and (20) in \citep{joachimssgt}. 
Hence the SGT is LS-KPCA with an RKHS defined by a graph based regulariser. Such regularisers have proven highly effective in SSL. A similar combination was proposed in \citep{chapslds} but with a non-convex gradient descent and more sophisticated loss functions. 

The RKHS derivation of SGT has various advantages. First, our experiments show that in some cases it can be more effective to use a normal Gaussian kernel RKHS regulariser rather than a graph based one. As we see in Section \ref{SECexperiments}, this happens particularly when the data density is adversarial in the sense of defying the so-called manifold assumption (that the data lie near a low dimensional sub-manifold of the input space \citep{chapslds}), in which case the graph based regulariser may be inappropriate. It is also straightforward to smoothly transition between the two regularisers as in \citep{belkinambient}. This transitioning can equivalently be obtained by simply including the graph Laplacian regulariser as an additional term in \eqref{EQNlskpcarwf1} of the  form 
$
\sum_{i,j} w_{ij} \lnb f(\xvec_i)-f(\xvec_j) \rnb^2,
$
since the overall problem can be converted just as easily as before to an optimisation problem in $\alphavec$ via the Lagrangian/representer theorem. 
Various other interesting options are straight-forward due to this flexibility. For example, problem specific invariances may be incorporated as in \citep{chapelleinvariance} by penalising loss terms which are functions of the gradient of $f$. This simply leads to an expansion for the optimal $f^\star$ which contains gradients of the kernel function \citep{representer}, and this leads immediately to finite dimensional optimisations similar to those in $\alphavec$ in our formulations.
Finally, compared with the graph cut derivation of the SGT our RKHS derivation permits a natural out of sample extension.

To complete our comparison with other methods let us finally mention LR-KPCA. This algorithm is a greater departure from previous work. It is related to logistic regression and of course LS-KPCA, but seems to be the first iterative algorithm to take advantage of the exact solution of LS-KPCA provided by \citep{gandalf} as part of an inner loop.

\section{Experiments and Discussion}
\label{SECexperiments}
\begin{table*}
\begin{center}
{\fontsize{7}{7.5}\selectfont \centering
\begin{tabular}{|c|c|c|c|c|c|c|}
\hline
& g241c & g241d & Digit1 & USPS & BCI & Text  \\
\hline
\hline
LR-KPCA & \bf{15.09} (4.57) & 49.71 (3.96) & \emph{15.52} (4.01)  & \emph{21.41} (3.06) & \emph{47.46} (1.75) & \emph{32.5} (3.57) \\
LS-KPCA & \bf{14.70} (1.83)     & 48.74 (3.91)  & \emph{13.86} (2.99)  & \emph{23.75} (3.20)  & \emph{48.44} (2.57)  & \emph{32.71} (3.00)   \\
MV-KPCA & \bf{14.12} (2.28)     & 50.24 (4.03)  & \emph{12.32} (3.69)  & 62.02 (12.70)  & \emph{50.04} (0.89)  & \emph{32.47} (3.35)   \\
MV-KPCA-10 & 36.34 (7.54)  & \emph{32.96} (8.56)  & 21.37 (4.69)  & 55.59 (3.92)  & \emph{48.46} (2.15)  & \emph{34.74} (7.88)   \\
KPCA-10 & 28.75 (4.26)     & \emph{32.98} (8.58)  & 21.36 (4.67)  & 35.27 (13.59)  & \emph{48.16} (2.91)  & 36.00 (9.76)   \\
\hline 
\hline
LR-KPCA & \bf{12.64} (0.46) &23.8 (3.00) & \emph{4.60} (1.07) & 11.35 (1.87) & \emph{32.25} (2.04) & 27.66 (2.88) \\
LS-KPCA & \bf{13.12} (0.45)     & 22.93 (3.19)  & \emph{4.08} (1.34)  & \emph{7.51} (1.01)  & \bf{29.03} (2.20)  & \emph{24.96} (2.61)   \\
MV-KPCA & \bf{12.82} (0.42)     & 49.95 (1.64)  & \emph{3.89} (1.08)  & 20.11 (5.61)  & 49.33 (1.59)  & 30.80 (2.03)   \\
MV-KPCA-10 & 17.78 (1.63)  & \emph{17.04} (2.70)  & 9.40 (1.59)  & 10.08 (1.82)  & 37.17 (3.76)  & 27.50 (1.57)   \\
KPCA-10 & 18.11 (1.85)     & 21.14 (3.06)  & 7.42 (1.41)  & 14.46 (1.53)  & 48.50 (1.71)  & 33.79 (3.59)   \\
\hline
\end{tabular}}
\end{center}
\caption{\emph{mean (std-dev)} errors over 12 splits for 10 (top half) and 100 (bottom) labeled points out of 1500, on the benchmark sets of \cite{sslbook}.
\label{TABLEbenchmark}}
\end{table*}

We tested on the six binary benchmark data sets of \citep{sslbook} as follows.
\subsection{Gaussian Kernel}
Each error in Table \ref{TABLEbenchmark} corresponds to a mean (standard deviation) over the twelve test splits
supplied with the data sets, for each of the two supplied cases: 10 (top half of table) and 100 (bottom half) labeled points.
We used the Gaussian kernel $k(\xvec,\yvec) = \exp(-\gamma\norm{\xvec-\yvec}^2)$, 
so for each split we had to choose the parameters $c$ (for LS- and LR-KPCA), $s$ (for LS-, LR- and MV-KPCA) and $\gamma$ (for all three and also KPCA).
To choose these parameters for each split, we performed 10 fold cross validation over the labeled points of that split.
This model selection procedure can be problematic, especially for the 10 labels case in which the cross validation estimate is especially unreliable,
but to be fair we always followed this procedure.
Italics in the table indicate signficantly best results amongst our algorithms, in the sense of having a mean error less than the mean error plus standard deviation of the method with lowest mean. Bold indicates best mean result over all published results in the study \citep{sslbook}.

The algorithms listed in Table \ref{TABLEbenchmark} are the following. LS- and MV-KPCA correspond to using the first eigenfunction from those problems as the classifying function. 
For MV-KPCA-10 and KPCA-10 however we took the top ten eigenfunctions, and used the values of these functions to train a hard margin, linear SVM (which sees these ten values on the labeled points only). Our motivation for testing MV-KPCA-10 was that by penalising within class variances rather than a signed class label loss term,
this algorithm may be flexible enough to extract multiple relevant features. We include KPCA-10 as a baseline for comparison.
The SVM was also applied to the first eigenfunction of LR-, LS- and MV-KPCA, although there the optimisation is trivial and merely sets a threshold.

Directly comparable error rates for eleven other algorithms are included in \citep{sslbook}, and the chapter with these numbers is freely available online.
It turns out that we obtain best mean performance compared with all methods on \emph{g241c} (for 10 and 100 labels) and for \emph{BCI} (for 100 labels only),
and generally competitive performance overall.
Comparing the different methods it turns out that one of the strongest competitors is the SGT \citep{joachimssgt}, which appears overall to be significantly better than our methods on these data sets. However, as explained in Section \ref{SECrelationship} the SGT is in fact a special case of our LS-KPCA. The main difference lies in the graph based regularisation of the SGT rather than our plain Gaussian kernel RKHS norm regularisation. Other more subtle differences include Joachims' choice of graph connectivity and weights, use of a non-trivial spectral renormalisation of the graph Laplacian, and rebalancing based on relative class frequencies \citep{joachimssgt}. Since these options are also possible for the more general LS-KPCA, we can argue that LS-KPCA should be attributed with the best performance of the published SGT results and our results here. Most importantly, it is clear that the most meaningful features of our results are hence captured in the relative performances of our different methods. However, one point we can make with regard to absolute performance measures, is that our best performance on \emph{g241c} (a sythetic data set composed as samples from two highly overlapping Gaussians, one per class) seems to indicate that LS- and MV-KPCA are able to handle non manifold like data effectively, presumably due to the non graph based regularisation.

Our algorithms do utilise the unlabeled examples, as we significantly out-perform the purely supervised baseline methods \citep{sslbook}. This is further evidenced by the fact that  KPCA-10 is never significantly better than the other variants, and often significantly worse. The mediocre performance of KPCA-10 on these datasets is surprising, as this algorithm seems reasonable for SSL, and was previously proposed for exactly that \citep{sskpcacrap}. We also found that LR-KPCA is rather similar to LS-KPCA. This does not seem to be due to computational problems since the iterative reweighting scheme always converged to high precision within 20 iterations. It may be due to the coarse grid we were forced to use for the cross validation parameter search of LR-KPCA, due to its being rather expensive to solve. It is expensive since each re-weighting step requires the LS-KPCA type solution, which itself requires of the order of 10 matrix inverses of size $m$ during the zero finding phase described in Section \ref{SECsecular}. Although expensive, a more refined search would be possible with sufficient computational resources, but would presumably only lead to modest improvement. Rather, it seems that the squared loss of LS-KPCA is reasonable for these problems, which agrees with the fact that a significant amount of work has been in precisely the opposite direction to our LS-KPCA $\rightarrow$ LR-KPCA. By this we mean the least squares SVM \citep{lssvm} where the hinge loss is actually replaced by a squared loss for classification (although the use of a squared classification loss is relatively uncommon overall in the literature). Moreover, in SSL the labeled loss term plays a diminished role in comparison to normal supervised learning as in the LS-SVM. Nonetheless, LR-KPCA performs strongly and intuitively on the two moons toy dataset as depicted in Figure \ref{FIGlrkpca}. 

\subsection{Combined Graph Diffusion and Gaussian Kernel}

The main difference between our LS-KPCA and the SGT \cite{joachimssgt} is the extra degree of freedom afforded the fact that LS-KPCA may utilise an arbitrary kernel function, rather than being restricted to a graph based representation. To demonstrate the value of this degree of freedom we now experiment with LS-KPCA using a convex combination of a graph diffusion and a normal Gaussian kernel, namely 
\begin{align} \nn
 K = w K_{\gamma} + (1-w) \exp(-\tau L),
\end{align}
where $K_{\gamma}$ is the kernel matrix associated with the Gaussian kernel as in the previous sub-section. $L$ is the normalised graph Laplacian
defined by $L = \diag{S \evec}-S$, 
where $S = \diag{W \evec}\mhalf W \diag{W \evec}\mhalf$. Here $W$ is the edge weight matrix for the graph. To construct $W$ we employed a standard nearest neighbour connectivity, and assigned Gaussian edge weights with respect to the pairwise Euclidean distance of connected vertices. We set the bandwith of this edge weight Gaussian to be equal to the mean squared pairwise distance between connected points, and removed self connections so that $W$ has zero entries on the main diagonal. Starting with $w=0$ (pure graph kernel), we chose the number of nearest neighbours, $\tau$, and the parameters $c$ and $s$ of \eqref{EQNlskpca1}-\eqref{EQNlskpca2} using a leave one out procedure which we accelerated by exploiting Cholesky up- and down-dates. This pure graph kernel based method is listed as \textit{G-LS} in Table \ref{TABLEnobold}. Given those optimal parameters, we then fixed $\tau$ and the number of nearest neighbours, and then selected $w$, $\gamma$, $c$ and $s$ again using leave one out, in order to asses the relative benefit of using a non graph based kernel in this experimental setting. The results for the case of mixing the Gaussian and diffusion kernels is listed in Table \ref{TABLEnobold} as \textit{M-LS}. We see that incorporating the Gaussian kernel in this manner never degrades the performance of the pure graph based algorithm, while for data sets \textit{g241c} and \textit{BCI} it significantly improves the performance.
\begin{table*}[t]
{\fontsize{7.5}{7.5}\selectfont
\begin{center}
\begin{tabular}{|c|c|c|c|c|c|c|c|}
\hline
 & Digit1 & USPS & BCI & g241c & COIL & g241d & Text \\ 
\hline \hline
\textit{G-LS} & 18.15 (7.79) & 25.37 (13.47) & 48.93 (2.33) & 42.29 (7.91) & 64.42 (5.21) & 46.81 (4.00) & 41.85 (7.08) \\
\textit{M-LS} & 15.95 (7.11) & 25.70 (11.38) & 48.93 (1.74) & 36.20 (13.70) & 73.58 (10.87) & 47.66 (3.80) & 39.92 (7.60) \\
\hline \hline
\textit{G-LS} & 2.70 (0.95) & 5.75 (1.28) & 48.06 (2.64) & 29.36 (6.19) & 25.33 (10.96) & 32.50 (6.06) & 26.67 (2.35) \\
\textit{M-LS} & 3.88 (3.00) & 5.70 (2.43) & 37.33 (7.93) & 16.63 (3.54) & 31.98 (23.72) & 24.02 (4.34) & 25.71 (2.04) \\
\hline
\end{tabular}
\end{center}
}
  \caption{\label{TABLEnobold} Mean and standard deviation percentage errors for 10 (top half of table) and 100 (bottom half) labeled points out of 1500. \textit{G-LS} is LS-KPCA with a graph diffusion kernel, while \textit{M-LS} is LS-KPCA with a combined Gaussian and graph diffusion kernel.}
  \end{table*}

\section{Conclusions}
\label{SECconclusions}

We have proposed three variants of KPCA for semi-supervised learning. All three are able to benefit from the unlabeled data, and lead to competitive overall results on benchmark sets. LS-KPCA generalises the powerful SGT algorithm \citep{joachimssgt}, thereby admitting various alternative algorithms due to the flexibility of the RKHS setting. Moreover, our RKHS based derivation of the more general case is conceptually cleaner than that of the SGT, which was originally derived from a relaxed spectral graph cut perspective. Both LS-KPCA and the SGT utilise \cite{gandalf} to obtain the globally optimal solution to their non-convex optimisation problems. We interpret this as a useful tool for problems related to that of the T-SVM, by considering the variance term in LS-KPCA as analogous to the T-SVM unabeled loss function. We also proposed the more sophisticated LR-KPCA, which implements a classification oriented sigmoid loss function via a reweighting scheme. This reweighting scheme also utilises the globally optimal solution of LS-KPCA in an inner loop. 

Generally speaking, we believe that the formulations in \citep{gandalf} are powerful, and perhaps under-utilised in machine learning. We hope to uncover a family of interesting algorithms (particularly for semi supervised learning) by studying re-weighted versions of these formulations. Here we presented an iterative re-weighting of the loss in \eqref{EQNlskpcarwf1} (as in LR-KPCA). Also interesting is the possibility of reweighting the summand in \eqref{EQNlskpcarwf2} in order to obtain more sophisticated unlabeled loss terms. This is more complex, and we plan to investigate this direction in the future.

\bibliographystyle{splncs}
\bibliography{walder}
\end{document}